\documentclass[runningheads,a4paper]{llncs}
\usepackage{wrapfig}
\usepackage{times}

\usepackage{graphicx}
\usepackage{mathtools}
\usepackage{xspace}
\usepackage{subfig}
\usepackage{enumitem}
\usepackage{xcolor,listings}
\usepackage{amssymb}
\usepackage{url}
\usepackage{cite}
\urldef{\mailsa}\path||%

\lstdefinestyle{WW}{
  basicstyle=\ttfamily\scriptsize,
  escapechar=@,
  numbers=none
}
\def\given{\mathrel{\vert}}
\def\variable#1{\text{\textsl{#1}}}

\DeclarePairedDelimiter\abs{\lvert}{\rvert}%
\DeclarePairedDelimiter\norm{\lVert}{\rVert}%
\makeatletter
\let\oldabs\abs
\def\abs{\@ifstar{\oldabs}{\oldabs*}}
\let\oldnorm\norm
\def\norm{\@ifstar{\oldnorm}{\oldnorm*}}
\makeatother

\makeatletter
\def\old@comma{,}
\catcode`\,=13
\def,{%
  \ifmmode%
    \old@comma\discretionary{}{}{}%
  \else%
    \old@comma%
  \fi%
}
\makeatother

\renewcommand{\dh}{\fontencoding{T1}\selectfont{\symbol{240}}}
\begin{document}

\title{Why Artificial Intelligence Needs a Task Theory \\ \vspace{0.2cm} \large --- And What It Might Look Like}
\titlerunning{Why AI Needs a Task Theory}  

\author{Kristinn R. Th\'orisson\textsuperscript{1,2,}%
\and Jordi Bieger\textsuperscript{1} \and Thr\"ostur Thorarensen\textsuperscript{1} \and \\ J\'ona S. Sigur\dh ard\'ottir\textsuperscript{2} \and Bas R. Steunebrink\textsuperscript{3}\\}

\institute{\textsuperscript{1} Center for Analysis \& Design of Intelligent Agents, Reykjavik University \\ \textsuperscript{2} Icelandic Institute for Intelligent Machines, Reykjavik\\
\textsuperscript{3} The Swiss AI Lab IDSIA, USI \& SUPSI \\
\mailsa\\} 

\authorrunning{Th\'orisson et al.} %
\tocauthor{Kristinn R. Th\'orisson, Jordi Bieger, Thr\"ostur Thorarenssen, Jona S. Sigurdard\'ottir, Bas Steunebrink}

\maketitle  %

\begin{abstract}
The concept of ``task" is at the core of artificial intelligence (AI): Tasks are used for training and evaluating AI systems, which are built in order to perform and automatize tasks we deem useful. 
In other fields of engineering theoretical foundations allow 
thorough evaluation of designs by methodical manipulation of well understood parameters with a known role and importance; this allows an aeronautics engineer, for instance, to systematically assess the effects of wind speed on an airplane's performance and stability. 
No framework exists in AI that allows this kind of methodical manipulation: Performance results on the few tasks in current use (cf.~board games, question-answering) cannot be easily compared, however similar or different. The issue is even more acute with respect to artificial \emph{general} intelligence systems, which must handle unanticipated tasks whose specifics cannot be known beforehand.
A \emph{task theory} would enable addressing tasks at the \emph{class} level, bypassing their specifics, %
providing the appropriate formalization and classification of tasks, environments, and their parameters, resulting in more rigorous ways of measuring, comparing, and evaluating intelligent behavior. Even modest improvements in this direction would surpass the current ad-hoc nature of machine learning and AI evaluation. Here we discuss the main elements of the argument for a task theory and present an outline of what it might look like for physical tasks.

\end{abstract}

\section{Introduction}
Artificial intelligence (AI) research is mostly aimed at the design of systems that can perform tasks that currently require human intelligence. 
AI systems interact with ``environments'' that contain all relevant existing objects and the rules by which they interact, while ``tasks'' assigned to an agent describe (un)desirable environment states that should be brought about or avoided.  We refer to the tuple of a task and the environment in which it must be accomplished as the ``task-environment''. 
Specialized AI systems are often made with a single task in mind, while systems aspiring to artificial \emph{general} intelligence (AGI) aim to tackle a wide range of tasks that are largely unknown at design time. 
Tasks can be divided in various ways into different sets of subtasks, and intelligent systems must make a choice about what tasks to pursue.  
Finally, tasks are used for the evaluation and training of these systems. 
So while the concept of ``task" is at the very core of AI, no general theory exists about their properties.\footnote{Classical planning has hierarchical task networks~\cite{georgievski_overview_2014}, but subtask decomposition is almost always done manually and there is no real analysis of tasks on a general level. Some people working on AI evaluation --- one of task theory's primary applications --- attempt to analyze some properties of task-environments, but they don't go beyond complexity and difficulty-related measures~\cite{hernandez-orallo_stochastic_2015}. } 
Tasks are generally selected on an ad-hoc, case-by-case basis without any deep understanding of their fundamental properties or how different tasks relate to each other. 

This is very different in many other fields. When for instance a new airplane needs to be designed to perform a certain family of tasks (say, move 300 passengers across the Atlantic within 8 hours)
aeronautics engineers employ theories firmly rooted in physics to turn  task parameters into (preliminary) requirements for the design of that airplane. 
They evaluate the design by running simulations, where parameters of the task and environment --- air pressure, wind speed, turbulence, humidity, precipitation, runway length, travel distance, etc. --- are changed to see their effect on the artifact's behavior. 
When an engineer changes an environmental variable (e.g.~wind speed) or feature of the task (e.g.~descent angle), its meaning is well understood in the context of an airplane's task-environment, which allows its resulting behavior to be readily understood. 
Evaluating the performance of complex systems usually involves the use of \emph{a battery} of such tests, each of which may be composed of a small or large set of atomic tasks. 
The thoroughness of evaluation depends on the diversity of the tests employed, chosen to provide a comprehensive picture of the system's behavior, in their ``comfort zone" as well as at the fringes of their target operating ranges. 
Finally, when the airplane is in use, decisions regarding its deployment are informed by its physical properties \emph{and} the task specification (e.g.~weather, range and cargo for a planned flight). 

In the absence of such theories for AI, people sometimes attempt to use theories from other domains. 
Extensive domain knowledge is often used in the design of (narrow) AI systems, but even here it is often a challenge to select the best machine learning techniques, gather the right knowledge, and optimize the system's performance. 
Experience in other domains tends to be unhelpful as different tasks cannot be compared or related to each other. 
Many researchers have turned to theories of human psychology and psychometrics to evaluate AI systems, resulting in human-centric tests (e.g.~the Turing Test, the Lovelace Tests, the Toy Box Problem, the Piaget-MacGuyver Room, and AGI Preschool --- see also the latest special issue of AI Magazine~\cite{marcus_beyond_2016}). %
Many of these tests are limited to providing a binary (and to this day universally negative) answer to the question of whether a system is truly intelligent. 

The range of possible AI systems and their capabilities is quite a bit greater than that of animals: 
Humans can't systematically reduce the size of their semantic memory, say, or replace their motor control scheme in an instant. 
Similarly, the range of tasks that AI systems might encounter is much greater than that of airplanes. 
AGI systems explicitly target diversity and complexity, aiming for a broad range of behaviors on a large set of complex tasks in a number of diverse environments. 
The range of evaluation tools the field has in its toolkit should reflect this breadth. 

Some evaluation methods have been proposed that cover a wider range of tasks, but they are either still too specific --- e.g.~general (video) game playing of a handful of manually created games~\cite{levine_general_2013,genesereth_general_2014} --- or they are so abstract that it is difficult to relate them to tasks in the real world --- e.g.~procedural generation of almost completely random tasks~\cite{legg_approximation_2011-1,garrett_tunable_2014}.\footnote{More discussion of various evaluation methods can be found in our previous publication~\cite{thorisson_towards_2015-1}.}
To adequately evaluate AGI-aspiring systems, rigorous tests profiling cognitive abilities such as transfer of training, knowledge retention, attentional control, and knowledge acquisition rate, would be highly valuable~\cite{thorisson_towards_2015-1}.
A deep and general understanding of tasks and the ability to construct and compare related tasks would greatly facilitate the design of an evaluation tool with these capabilities.

Ideally we would have good theories for all aspects of intelligence/competence/skills assessment, analysis and development. No classification scheme or architectural principles exist at present that seem likely to provide a unifying framework for research on (artificial) intelligences, and no such framework seems likely to spring forward in the near future, as researchers in the field don't even agree on which aspects are necessary components for a system to be called intelligent. A formal approach to tasks and environments can, however, begin to be undertaken, as they %
offer readily measurable physical features. %
Such a theory could take many forms, but rather than rooting it in computer science, the early results of the field of AI, or in human task analysis~\cite{robinson_task_2001}, we think it important to develop a task theory grounded in the physical realm.%

Note that we are not targeting a theory of learning, a theory of agents, or a theory of evaluation --- quite the contrary, we want to focus exclusively on the task-environment so as not to mix these; this may include the agent's body, but not its controller (``mind"). Otherwise we run the risk of continuing the conflation of the learner with its task and environment. (This aim, however, highlights the need for a proper theory of learning, agents and pedagogy --- these are not mutually exclusive but in fact ultimately necessary for achieving comprehensive evaluation of a wide range of learners in a wide range of circumstances.) Note also that we are not proposing to represent all tasks and environments accurately or in every detail, but rather to develop a task-environment representation that suffices for modeling a wide range of (the most important aspects of) these, in a way amenable for simulation, and possibly analytical computation.

In the rest of this paper we look at the requirements for a task theory of this kind, its potential applications, and outline some examples for what it might look like as a way to further clarify our intent and provide a case for its potential. 
\section{What We Might Want From a Task Theory}

Three important aspects of AI research include its \emph{evaluation, pedagogy} (training/learning), and \emph{design}. 
An AI system is designed for a particular role, involving performing a task or range of tasks. 
Tasks are also at the core of their pedagogy and evaluation. A properly conceived task theory could help with all these aspects of AI research.

\paragraph{Evaluation}

Evaluation of AI systems allows researchers to find the strengths and weaknesses of their creations and measure progress through comparisons with earlier versions and other systems. 
Evaluation of general intelligence is complicated by the facts that 1) no fully functional AGI systems exist to date, 2) different systems have different non-overlapping capabilities, and 3) tests must measure (progress towards) some general cognitive ability rather than performance on a specialized task~\cite{hernandez-orallo_ai_2014}.

When an agent, whether artificial or natural, is assigned a task, three things may be in flux: (a) The task --- 
which prescribes what goals should be achieved and/or what situations should be avoided%
, (b) the environment in which the task is performed --- which may act independently of the agent, and (c) the agent itself --- whose perception, memory, goal structures, and other cognitive features are affected by the task and environment. %
To analyze the effects of changes in one element of an interacting system, variation in the others must typically be controlled to prevent contamination of results. 
To quantify the performance of a system, or compare any set of systems, the task and the environment must be held constant, %
otherwise what is measured cannot be reliably attributed to the agent's performance. %
This would suggest that AI systems can only be compared if they are evaluated on the same task-environments. 
AI researchers would need to settle on a standardized set of tests to be administered (unmodified!) to all AI systems that we want to compare. 
However, it is unreasonable to assume that all of these systems will be, or even \emph{can} be, tested on exactly the same task(s). 
This would require a test that (a) can be used for all AI systems and (b) is discriminatory between all pairs compared --- clearly an impossibility. 
Even if the whole field settled on a standardized test battery today, it would likely become obsolete as AI systems evolve, as well as fall victim to specialists on the test: the very antithesis of AGI. 
We must accept that different systems are, and will continue to be, developed with different philosophies in mind, as this is how research is done.
Until we have systems with full generality (but possibly not even then) we need methods for evaluating intermediate milestones on different paths towards AGI. %
Given results from two different AGI-aspiring systems on two different tasks, we could compare them if we had a way of relating the tasks to each other on (some or all) key dimensions. 
This is precisely what a task theory would enable, thus removing the need for a standardized test battery applicable to all AI systems. %
All systems --- no matter how simple or advanced --- could be evaluated in their ``natural habitat'' and still be compared to each other, the quality of which would be determined in part by the power of the theory. 

We've argued before that a good A(G)I evaluation framework should enable the easy manual and automatic construction of task-environments and their variants as well as facilitate the analysis of parameters of interest~\cite{thorisson_towards_2015-1}. 
A task theory should similarly allow us to relate the features of a task to measurable physical and/or conceptual aspects, enabling comparison of similar and dissimilar tasks, and facilitate the construction of task-environments and variations on known tasks without changing their \emph{nature}, so that we may select or design tasks capable of measuring various aspects of AIs.%
We would like to be able to compose and decompose tasks and environments, and scale them up or down in size or complexity in accordance with robust and well-understood principles. 
In order to provide a characterization of task-environments, measures ought to be defined for properties like determinism, ergodicity, continuity, asynchronicity, dynamism, observability, controllability,%
periodicity, and repeatability~\cite{thorisson_towards_2015-1}.

\paragraph{Pedagogy} Learning systems must be trained for the task(s) that they are created for. 
In some (simple) cases such systems may be able to learn everything on their own, but even when teaching isn't strictly necessary it can improve the training, e.g.~by speeding it up~\cite{bieger_raising_2014-1}.
Teaching is done interactively using various forms of communication, demonstration and/or by assigning carefully selected or constructed tasks to the student system so that it may learn the relevant knowledge and skills faster --- e.g.~teaching simplified versions of a task's component parts before teaching the whole. 
Task theory should help in the construction of appropriate training scenarios and task features. 
A deeper understanding of tasks would also enable systematic use of analogies and abstraction that can be explained to a student system. %

\paragraph{Design} %
A task theory could also help alleviate some of the trial-and-error involved in designing AI systems for a particular set of tasks or task types. %
Currently designers build up informal, difficult to verbalize experience and intuition for matching certain system features with their understanding of the task at hand. 
A task theory would allow for systematic characterization and comparison of different tasks, and thus take out some or significant parts of that guesswork. %
While full-fledged testing and evaluation will remain a necessity, the ability to e.g.~predict time and resource use for a task would create a rapid feedback loop for designing the AI's body --- and if combined with a theory of learning systems, also its controller.

\section{Requirements for a Task Theory}

A task theory should cover all aspects of tasks and the environments in which they must be executed; in short, it should enable us to model tasks in a way that supports:

\begin{enumerate}
	\item \emph{Comparison} of similar and dissimilar tasks.
	\item \emph{Abstraction} and \emph{concretization} of (composite) tasks and task elements.
	\item Estimation of time, energy, cost of errors, and other resource requirements (and yields) for \emph{task completion}. 
	\item Characterization of task complexity in terms of (emergent) quantitative measures like \emph{observability}, \emph{feedback latency}, form and nature of \emph{information/instruction} provided to a performer, etc. %
	\item Decomposition of tasks into subtasks and their atomic elements.
	\item Construction of new tasks based on combination, variation and specifications.
\end{enumerate}

These requirements should enable 
the applications mentioned in the last section. A computational task theory would provide a foundation for frameworks/toolkits that can \emph{simulate} a wide variety of tasks in the form of \emph{task models} constructed according to the theory, automatically \emph{produce variants} of tasks over some desirable distributions, and run evaluation tests in \emph{batch mode} to provide a vast amount of performance data for any set of controllers and AI systems.
Estimation of time, energy and other resource requirements (and yields) for task completion can be used to design effective and efficient agent bodies, judge an agent based on comparative performance, and make a cost-benefit analysis for deciding what (sub)tasks to pursue. 
The models constructed according to this theory are used to estimate the AI's ability to perform real life tasks if provided with the actuators and sensors contained within the model.\footnote{Recall that a task theory would include the limitations that the body of an agent imposes --- its interface to the task-environment.} The performance of the AI can be described by the energy as a function of time and the precision at which the task was completed, where the highest possible attainable precision is defined by the laws of physics.  

Performing a task in the real world requires time, energy and possibly other resources such as money, materials, or manpower. 
Omitting these variables from the task model is tantamount to making the untenable assumption that these resources are infinite~\cite{wang_assumptions_2011}. 
Any action, perception and deliberation must take up at least some time and energy. %
Therefore every task that we model must have these components. 

Characterization of tasks can facilitate their comparison at a high level, enabling us to contextualize the performance of different systems on different tasks. It would also allow us to correlate these quantitative measures with the performance of a particular agent in order to seek out tasks that are more suitable to it (e.g.~for evaluation or teaching). 
Decomposition, abstraction and comparison all facilitate a deeper understanding of task-environments that could potentially be communicated to a student. %
Comparison of tasks in terms of e.g.~environment contents and structure, can have similar benefits to comparison in terms of high-level measures, but additionally help with analogical reasoning and transfer of knowledge. 

Importantly, these features would facilitate the construction of new tasks: they provide building blocks (decomposition), information about fundamental features to keep (abstraction) and a way to ensure a certain amount of similarity (comparison). Task theory should allow for the construction of variants of tasks with differing levels of similarity, and even support generation according to a high level specification of the characteristics under item 4. Such constructions would allow for tailor-made training environments and evaluation tools that support a wide range of systems.

\section{What a Task Theory Might Look Like}

Completion of a fully-fledged task theory that meets all of our requirements is a rather large endeavor that has only just begun, and the requirements a task theory as described above can likely be met in a variety of ways. To make more tangible our aims with this work we outline now ideas for a concrete direction we are exploring for physical tasks, as a way to both ground the preceding discussion and provide some potential demonstrations of its feasibility. For this purpose we simulate a fully specified task model, allowing precise analysis. %
Producing definitions of key concepts compatible with physics is important to us, as we are looking for a theory that allows engineering of task-environments with measurable physical properties.%

Estimating the time and energy requirements of a compound task precisely for an arbitrary agent is tricky: We cannot simply expect that the agent will immediately pick an optimal action sequence to complete the task using the minimum amount of time or energy (different strategies may result in the optimization of time vs. energy). One approach would be to map out spaces of solutions and all possible action sequences. The ratio of these two will tell us the potential of a controller to fail the task --- more formally, the probability of its successful completion of a random performer. Having a number of dimensions for which to measure this ratio, including constraints of time, energy, and other factors, would mean that tasks could be profiled and positioned in a multidimensional manifold. By grounding such measures in absolute physical terms, distances between any two tasks would represent real physical measurements and tell us a lot about how the solution spaces for them compare. 

When done naively, mapping out the entire space of possible action sequences is only feasible for very small tasks. However, for compound tasks we could combine those small tasks in various ways to get larger tasks whose properties we can estimate (e.g.~serial composition multiplies the ratios). Alternatively, decomposition could be applied until the component parts resemble tasks whose properties we can calculate. 

In order to make any calculations or analysis however, we need a more concrete idea of fundamental concepts like \emph{environment}, \emph{state}, \emph{agent}, \emph{goal}, \emph{problem}, and \emph{task}. These concepts must be defined in a way to facilitate (modular) construction and analysis while preferably not straying too far from their intuitive notions.

\subsubsection{Environment} 
The highest level in our conceptualization of task-environments is a world $W$, which is an interactive system consisting of a set of variables $V$, dynamics functions $F$, an initial state $S_0$, domains $D$ of possible values for those variables, and a possibly empty set of invariant relations between the variables $R$: $W = \left\langle V, F, S_0, D, R\right\rangle$. %
The variables \(V = \left\{v_1, v_2, \dots, v_{\lVert V\rVert}\right\}\) represent all the things that may change or hold a particular value in the world. 
A system's dynamics can intuitively be thought of as its ``laws of nature". %
As a whole, the dynamics may be viewed as an automatically executed function that periodically or continually transforms the world's current state into the next: \(S_{t+\delta} = \operatorname{F}(S_t)\).  %
However, in practice it is often useful to decompose the dynamics into a set of transition functions: $F = \left\{f_1, f_2 \dots f_n\right\}$ where \(f_i: S^- \rightarrow S^- \) and \(S^-\) is a partially specified state.

Each variable $v$ may take on any value from the associated domain $d_v \in D$. 
For physical domains we can take the domain of each variable to be a subset of the real numbers. %
Invariant relations $R$ are Boolean functions over variables that hold true in any state that the system will ever find itself in. 
In a closed system (with no outside influences) the domains and invariant relations are implicitly fully determined by $F$ and $S_0$. 
In an open system --- where change may be caused externally --- explicit definition of domains and relations can be used to restrict the range of possible interactions. 

Environments are views or perspectives on the world. 
In their simplest form they can be characterized as slices or subspaces of the world, where all variables can take on a subset of the world's variables, each variable's domain is a subset of that variable's domain in the world, and only the relevant dynamics and invariants are inherited. 
Environments can be defined for different purposes (e.g.~for different tasks or agents), and their overlap and similarities can be analyzed. 
A task-environment should include all aspects of the world that are relevant to the completion of the task.

\subsubsection{State} 
A \emph{concrete state} \(S\) is a value assignment to all variables of a system: \(S = \bigcup_{v \in V} \left\{ \left\langle v, x_v \given x_v \in d_v \right\rangle \right\}\). 
A state is valid if and only if all relations hold true:\\ \(\operatorname{valid}(S) \iff \forall_{r \in R} r(S)\). 
A \emph{partial state} \(S^-\) only assigns concrete values to a subset of the variables in a system. %
For real variables partial states can be represented by using error bounds: \(S^- = \bigcup_{v \in V^-} \left\{ \left\langle v, x_l, x_u \given x_l < x_u \wedge (x_l, x_u) \subseteq d_v \right\rangle \right\}\). 
As such, a partial state really covers a set of concrete states. 
This concept is more practical, since it is rare that we precisely know or care about the value of every last variable: in most cases only a subset matters, and noise and partial observability make it impossible to know most values with absolute precision. 

\subsubsection{Agent} 
An agent $A$ is an embodied system consisting of a controller $C$ (the AI system) and a body $B$. The body is the agent's interface to the world and communicates signals from sensors to the controller, which in turn sends back commands to be turned into atomic actions by the body's actuators. %
The internals of the controller are beyond the scope of a physical task theory, and since any physical system is naturally embedded in the world, the body merely contains two lists of environment variables that the controller can directly read from and write to: $B = \left\langle V_S, V_A \right\rangle$. 
In other words: all sensors and actuators must be (physical) objects in the world. 

\subsubsection{Problem \& Goal} 
A \emph{goal state} $g$ is a desirable (partial) state that the agent should reach. 
A \emph{failure state} $\overline{g}$ is an undesirable (partial) state that the agent should avoid. 
An \emph{atomic problem} is specified by an initial state, goal states and failure states. %
\emph{Compound problems} can be created by operations like conjunction, disjunction and negation. 
A \emph{solution} is a sequence of (atomic) actions that results in a \emph{path} through the state space that reaches all of the goal states and none of the failure states. %
A problem for which a solution is known to exist is called a \emph{closed problem}.

\subsubsection{Task} 

Finally, a \emph{task} is a problem \emph{assigned} to an agent. %
This assignment includes the manner in which the task/problem is communicated to the agent --- e.g.~whether the agent gets a description of the task a priori (as in AI planning), receives additional hints, or only gets incremental reinforcement as certain states are reached. 
A task is performed successfully once the world's history contains a path that solved the problem. 

We could additionally attach utility functions to problems to measure the degree of success --- rather than just success / failure / in progress --- but we could also emulate this by assigning the agent multiple simultaneous tasks and the meta-goal of performing as many as possible. 
For instance, if a task is considered successful when a certain (partial) state is reached before time $t=2$, but it's even more desirable to do it in less time, then we could assign an additional task that only succeeds if the goal is reached before time $t=1$.

Two kinds of tasks are typically identified (cf.~\cite{wooldridge_introduction_2009}): what might be called \emph{achievement tasks} (e.g. ``ensure $X \approx G_X$ or $X \not\approx \overline{G_X}$ before time $t \geq 5$'') and \emph{maintenance tasks} (e.g. ``maintain $X \approx G_X$ or $X \not\approx \overline{G_X}$ until $t=10$),\footnote{We use approximate rather than precise equivalence between $X$ and its goal value $G_X$ because we intend for our theory to describe real-world task-environments, which always must come with error bounds.} where $X \in V$ and $G_X \in d_X$. Combinations are possible (``ensure $X \approx G_X$ between time $t = 5$ and $t = 10$''). 
Performing a task in the real world requires time, energy, and possibly other resources (money, materials, manpower). 
But taking physical constraints into account makes it clear that any goal state must be held (maintained) for a non-zero duration (at a minimum sufficiently long for the achievement to be detected). What seems like two kinds of tasks is thus actually just one kind of goal with particular parameter settings (an accomplishment goal is simply a goal whose state may be held for a short period of time, relative to the time it takes to perform the task which it is part of --- a maintenance goal is held for relatively longer periods of time). 
The highest attainable precision of a goal state is defined by the laws of physics and the resolution of sensors and actuators. 

For any human-level task in the physical world, even seemingly simple ones such as doing the dishes or going to the store to buy bread, $V$ and $F$ will generally be quite large.

\subsubsection{Example}

Consider as an example a simplified driving task modeled in task theory. The agent must drive over a frictionless surface towards a target that is some distance away by using the gas pedal to control the power (i.e.~the rate at which fuel is burned). The world might be initialized to 
$S_0 = \{\langle\variable{time}, 0\rangle, \langle\variable{energy}, 10\rangle, \langle\variable{position}, 2\rangle, \langle\variable{velocity}, 0\rangle,  \langle\variable{power}, 0\rangle, \langle\variable{mass}, 2\rangle%
\}$. 
The goal could be $g = \{\langle\variable{position}, > 10\rangle, \langle\variable{time}, < 5\rangle, \langle\variable{energy}, > 0\rangle\}$: every realistic task should have a deadline and energy budget. The dynamics are defined by using basic Newtonian physics:
`$\variable{time} \;\leftarrow\; \variable{time} + \delta$', 
`$\variable{energy} \;\leftarrow\; \variable{energy} - \delta\cdot\variable{power}$', 
`$\variable{position} \;\leftarrow\; \max(0, \variable{position} + \delta\cdot\variable{velocity})$', and 
`$\variable{velocity} \;\leftarrow\; \sqrt{\frac{2\delta\cdot\variable{power}}{\variable{mass}} + \variable{velocity}^2}$'.
The body might simply be defined by $B = \left\langle \{\variable{position}\}, \{\variable{power}\} \right\rangle$, meaning that the agent can observe the position and control the force. From the dynamics and initial values we can tell for instance that the time and position will never be negative. However, since the force can be externally controlled, we need to specify its domain: $d_\variable{power} = [0,10]$. 
The agent can solve the task in many ways, but the fastest is to use maximum power (this takes 2.863 seconds, but energy runs out after 1). %
We could perform a different analysis for optimal energy usage (e.g.~using 0.15 Joule/second solves the task in 9.865 seconds, leaving 8.52 Joules). 

More complex variants of this simple task can be made with some slight adjustments. %
Tasks from the same family are similar in nature and share many fundamental properties; they can be closely related, e.g.~if they only (slightly) differ in their initial states or allocated resource budget, and more distantly if few features are shared, such as tennis and football. 
Allowing the agent to choose which direction to move in, for example, increases the chance of the agent missing the target. Adding friction, wind, obstacles and hills will increase the complexity of the original task without changing the nature of the model itself. Another important way to make variations is by changing the resolution, noise and latency of sensors and actuators. The task can also easily be extended by adding clauses to the goal (e.g.~require that velocity becomes 0), adding more goals (e.g.~to move back to the start), or by adding a second dimension.

\section{Conclusions}
We have argued for the importance of a task theory for various aspects of AI research, highlighting system design, pedagogy, and evaluation. 
Such a task theory should allow for (1) estimation of time, energy and other resource requirements (and yields) for task completion, (2) characterization of tasks in terms of emergent quantitative measures like complexity, observability, etc., (3) decomposition of tasks into subtasks and their atomic elements, (4) abstraction of (composite) task elements, (5) comparison of similar and dissimilar tasks, and (6) construction of new tasks based on combination, variation and specifications. 
A physical task theory contains the specification of an agent's body, but not its mind, and by virtue of being rooted in the physical world (all worthwhile activities of AI systems will eventually result in physical events) time and energy must always be taken into account. 
A theory like this does not exist yet, and will need to be constructed piece by piece. The ideas presented here are our thoughts on how to start developing such a theory.

\subsubsection*{Acknowledgments} The authors would like to thank Eric Nivel for insightful comments. This work was sponsored by the School of Computer Science at Reykjavik University, by a Centers of Excellence Grant (IIIM) from the Science \& Technology Policy Council of Iceland, and by a grant from the Future of Life Institute.

\bibliographystyle{splncs03}
\bibliography{task_theory_references}

\end{document}